# Paying Attention to Descriptions Generated by Image Captioning Models

Hamed R. Tavakoli[†]    Rakshith Shetty[⋆]    Ali Borji[‡]    Jorma Laaksonen[†]

[†]Dept. of Computer Science, Aalto University, Finland.
[⋆]Max Planck Institute for Informatics, Saarbrücken, Germany.
[‡]Dept. of Computer Science, University of Central Florida, Orlando, USA.

## Abstract

*To bridge the gap between humans and machines in image understanding and describing, we need further insight into how people describe a perceived scene. In this paper, we study the agreement between bottom-up saliency-based visual attention and object referrals in scene description constructs. We investigate the properties of human-written descriptions and machine-generated ones. We then propose a saliency-boosted image captioning model in order to investigate benefits from low-level cues in language models. We learn that (1) humans mention more salient objects earlier than less salient ones in their descriptions, (2) the better a captioning model performs, the better attention agreement it has with human descriptions, (3) the proposed saliency-boosted model, compared to its baseline form, does not improve significantly on the MS COCO database, indicating explicit bottom-up boosting does not help when the task is well learnt and tuned on a data, (4) a better generalization is, however, observed for the saliency-boosted model on unseen data.*

## 1. Introduction

The recent advancements in machine learning, together with the increase in the available computational power, have increased the interest in solving high-level problems such as image-captioning [14, 61, 9, 13], scene and video understanding [51, 52, 39], and visual question answering [37, 2]. The main goal of these problems is an inference which ends in a human-like response. The nature of such responses often necessitates interaction between several low-level cognitive tasks, e.g., perception and sentence planning in describing images. Measuring the capability of a machine in replicating such interactions is challenging. Although the trivial assessment techniques facilitate understanding the average performance of the algorithms, we yet need more detailed studies to understand specific properties of the existing methods in comparison with a human baseline. In the domain of image description by machines, agreement with visual attention is one such case.

There exists various theories about human sentence construction and formation. Wundt's theory of sentence production motivates the role of object's importance in sentence production. He proposes that, in a free word positioning scenario, not bound by any traditional rule, the words follow each other according to the degree of emphasis on the concepts [56]. This theory implicitly motivates the role of what is later on recognized as saliency. Griffin and Bock [21] found some empirical supporting evidence by showing that while describing scenes, speakers look at an object before naming it within their description. Besides, there exists numerous studies which have utilized attention to analyze human sentence planning and construction in different scenarios including scene description. Most of their findings provide supporting evidence that the sentence formation and attention correlate [6, 40, 22, 26, 64]. Encouraged enough, we lay the foundation of this study on the role of saliency in the construct of image descriptions, where the order of named objects is momentous in a sentence.

**Contribution:** In this paper, we address two intriguing questions: (1) *How well do image descriptions, by humans or models, on a scene agree with saliency?*, (2) *Can saliency benefit image captioning by machine?* Answering the questions, we learn not only about the role of attention in describing images, but also about the quality of human-written descriptions and machine-generated ones. We first study the textual statistics of the sentences by human and machine. Then we investigate the attention correlation in the structure of human-written and machine-generated descriptions. To further evaluate the contribution of low-level cues, we propose a saliency-boosted captioning model and compare it against a set of baseline captioning models.

## 2. Related Work

**Image description generation.** There exists a wide range of captioning methods and models. They can be categorized into retrieval-based [19, 44, 25, 33], sentence generation [34, 16], and the models which combine the two paradigms [61, 18, 15]. In-depth study of these different models is beyond the scope of this article and falls within the surveys such as [5]. We, however, briefly address the four models utilized in this study. All the four models follow the popular encoder–decoder approach to captioning, wherein the encoder converts the input image to a fixed size feature vector and the decoder is the language model which takes the feature vector as input to generate a caption. "Neural Talk" [32] utilizes a vanilla recurrent neural architecture for the language model while using CNNs for encoding. "Microsoft" [18] employs multiple instance learning to learn a visual detector for words in order to utilize them in an exponential language model for sentence generation. The "Google" method [61] is a generative deep model based on recurrent architectures, more specifically *long short-term memory* (LSTM) networks. "Aalto" [52] employs object detection to augment features in order to boost the results in a framework similar to "Neural Talk", utilizing LSTM networks.

**Automated metrics of description evaluation.** The community often favors automated metrics over human evaluation due to their reduced cost, faster processing speed, and replicability. The current popular metrics of evaluation are mostly borrowed from or inspired by machine translation. Some of these metrics are *BiLingual Evaluation Understudy* (BLEU) [45], which signifies the precision and neglects recall, *Recall Oriented Understudy of Gisting Evaluation* for the *Longest* common subsequence (ROUGE-L) [35], which is based on the statistics of the sentence level structure similarities, *Metric for Evaluation of Translation with Explicit Ordering* (METEOR) [12] and *Consensus-based Image Description Evaluation* (CIDEr) [60]. Adopting an explicit word-to-word matching, METEOR addresses the weakness of BLEU caused by the lack of recall information. Recently, CIDEr was developed for image description evaluation. It is a similarity-based metric that computes the similarity of sentences by the occurrence of *n-grams*.

**Attention and language studies.** The joint study of attention and language covers different perspectives either to understand the language development process [21, 41] or to investigate the role of language in scene understanding and comprehension [48, 31]. There exists numerous research in this area and surveying all of them is beyond the extent of this article. Instead, we focus on some of the most relevant ones. It has been demonstrated that the eye gaze and object descriptions highly correlate [65]. Further, in-depth analysis of gaze behavior in scene understanding and description reveals that people are often describing what they looked at [64], promoting the notion of importance. In [4], the importance of objects is studied in terms of their descriptions where object referral indicates the object's importance. On the other hand, obeying natural scene statistics, the importance and saliency of an object are equal [68]. The saliency, in the form of bottom-up attention, is reported to act as a facilitator whereby salient objects are more likely to be reported in scene descriptions [28]. The role of perception and attention is, however, more than the decisive role of referral and can even influence the order of mentioned objects [11]. Thus, we employ attention to analyze image descriptions written by humans and generated by machines.

The attention and language studies are affected by the difficulties of relating visual information to linguistic semantics. To date, most of the attention and language studies often use object bounding boxes, which introduces a degree of inaccuracy, in order to identify attention on objects. As an alternative to bounding boxes, [20] employed precise hand-labelled object masks to investigate the relation between objects and the scene context. Nonetheless, such annotations are often avoided due to cost. Thus, Zitnick *et al.* [67] proposed using abstract images in conjunction with written descriptions for semantic scene analysis, which is impossible for natural images. In this paper, we follow a procedure similar to [20] and rely on precise hand-labelled object masks for natural images.

**How are we different?** It is worth noting that what distinguishes the present work from aforementioned works like [4, 64, 65, 28] is that we consider the machine generaed sentences in conjunction with the human written ones, enabaling us to compare machine and human. Furthermore, building on top of the findings of such comparison, we study the contribution of saliency in image captioning models.

## 3. Data

**Human descriptions.** Fig. 1 depicts examples of the data (images and their human-provided annotations) used in this study. There exist several famous datasets for the task of image description generation. At the time, the most popular dataset is the MS COCO [36]. It consists of over 200K images with at least 5 human-written sentences per image. Among large datasets, there exists Flicker8K [25] and its extention Flicker30K [63]. One of the earliest well-recognized datasets is UIUC PASCAL sentences [50]. It consists of 1K images selected from the PASCAL-VOC dataset [17] and 5 human-written sentences for each. The same image set is used in PASCAL-50S [60], where 50 human-written sentences are provided. The use of the PASCAL-VOC images gives PASCAL-50S the advantage of having rich contextual annotation information [43]. Furthermore, the same image set is used by [65] for gaze-based analysis of objects and descriptions, where the gaze is recorded during free-viewing separate from image descriptions. Combining the three data sets, i.e. [60, 65, 43], results in 1K images with 50 sentences, 222 precisely labelled ob-ject class categories, and gaze information. We call the new data as *augmented PASCAL-50S* (see Fig. 1).

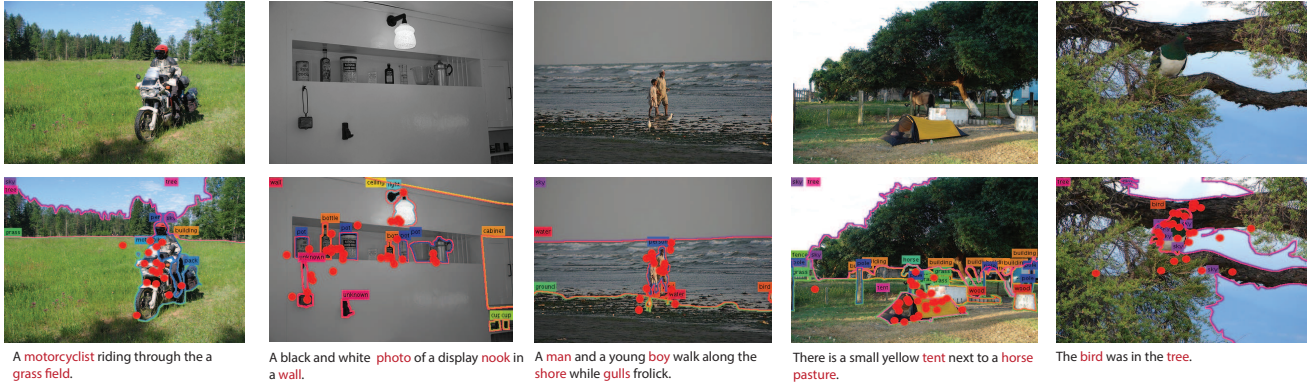

Figure 1. Example of the data used in this study, after merging the information from different sources. One reference sentence provided (with nouns highlighted), the boundary of each object is color coded, and the fixations are overlaid.

**Machine-generated descriptions.** The machine descriptions are generated by four captioning models, including: "Neural Talk" [32], "Aalto" [52], "Google" [61], and "Microsoft" [18]. All the models were trained on the MS COCO data set [9] and generated descriptions for the image set of augmented PASCAL-50S. It is worth noting that the models are not necessarily the same as those reported on the COCO leader board information page [1] since the scores are updated by new submissions. Having machine-generated descriptions, we analyze the differences between machine-generated sentences and human-written ones.

**Preprocessing.** To exploit the full potential of the data, we conduct a preprocessing step and compute a visual object category to sentence's noun (VOS) mapping. VOS mapping is a key ingredient for analyzing sentence constructs in terms of attention. It associates the object categories and their corresponding hand-labelled masks with the nouns in the descriptions.

The database consist of images with 222 unique class categories like, 'person', 'airplane', etc. They are accompanied with hand-labelled object masks, useful to establish a visual object to description mapping. To obtain such a mapping, we first identified all the nouns in the sentences by running a part of speech (POS) tagging software [38]. All the unique nouns were extracted, which accounts for 3760 nouns. We listed the top 200 most similar (similarity score > 0.18) nouns to each object class label using word2vec [42], trained on approximately 100 billion words from Google[1]. Then, we manually checked the correspondence between the listed nouns and the category labels to establish a mapping from the visual domain to noun descriptions. During this process, we also considered minor issues such as misspellings and identified the synonyms. The synonyms are manually identified by considering the relation to object class categories and the corresponding images by a human.

## 4. Analyzing Human Sentences

**The quality of descriptions.** The descriptions are written in a somewhat free from style. We, thus, studied their quality to gain a better understanding prior to any comparison with machine-generated descriptions. To this end, we looked into some factors, including correct syntactic grammar, the presence of a verb, and active and passive structures. The grammar checking was performed using a link grammar checking syntactic parser [55]. This process leaves 29646 sentences out of 50K (approx. 60%), that are not affected by grammatical errors. It is worth noting that we adopted conservative settings in the parser, which means the correct sentences can be slightly more than what is reported here. Among the grammatically correct sentences, using [38], we identified that only 19126 sentences (64%) have a verb, of which 17362 (90%) are active and 1764 (10%) are passive.

It is worth noting that applying the same procedure to the machine generated sentences, we learn that the sentences are all in active voice. To understand the reason, by analyzing the training data, i.e., the MS COCO, we notice that the models are presumably not exposed to enough passive sentences during training, and are incapable of generating such sentences. Consequently, we report the comparative analysis between the machine-generated descriptions and active human-written sentences.

**Object noun statistics.** Based on the information of VOS mapping for human descriptions, Fig. 2 summarizes the object noun statistics. On average 4.76 synonym terms are used to refer to one object class category (maximum 114). Some class categories have unusually high number of synonym terms as they can often be referred by specific type attributes, e.g., a 'person' can be identified as a 'boy', 'girl', 'man', etc. The top 20 words with the maximum number of synonyms are depicted in Fig. 2. The probability of object referrals in terms of noun orders is computed between different groups of humans. Two random groups of human descriptions are selected (10 and 40) and the probability of the order of a noun is computed, revealing that people tend to refer to objects more often at the beginning of the sentences. There is also a tendency for naming more objects when the image is more populated, except when the image is overcrowded (see the scatter plot in Fig. 2).

---

[1] https://code.google.com/archive/p/word2vec/

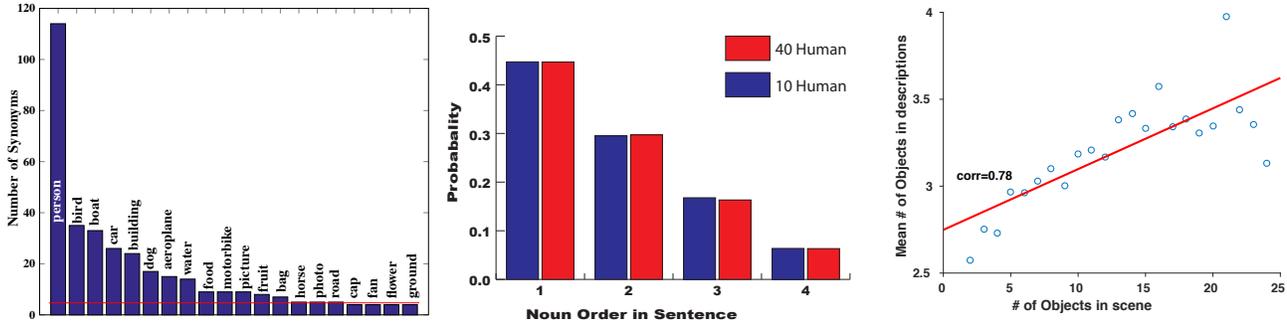

Figure 2. Analyzing human sentences: top 20 class categories with maximum number of synonyms, the probability of a noun referral with respect to the order of appearance in a sentence, and the correlation of number of objects in descriptions and scene.

**Object importance.** We measured the object importance in terms of object size and attention. We computed the *normalized object size*, defined as $nos =$ area of object/area of image, where the object area is obtained from the annotation mask. Then, the average over all object instances in all images is reported as the mean normalized object size. To measure the attention, knowing the mask of an object, we used the amount of fixations overlapping with the mask:

$$attention = \frac{\text{\# of fixations on the object}}{\text{\# of fixations on the image}}. \quad (1)$$

We reported the mean of attention over all the instances of an object class category. Fig. 3 summarizes these statistics, signifying that some objects are often more attended in agreement with the findings of [65], providing us some idea about the importance and saliency of objects. We then computed the visual occurrence probability of an object with its description-based occurrence. Overall, highly-attended objects are more probable to be referenced in the descriptions, e.g., 'person'. There are, however, some exceptions that are objects which are rarely referenced explicitly and still have high attention value, e.g., 'bird cage'.

## 5. Machine vs. Human

**Object importance and its referrals.** We compute object size and attention as a function of referral order to study the importance of objects in conjunction with their referrals in the descriptions. For each image and description pair, we first identify the annotation masks of the described objects using the VOS mapping. Afterwards, the object size and attention are measured using the object annotation masks while considering the order of nouns in descriptions. Going though all the description and image pairs, the mean normalized object size ($nos$) for each noun order and average *attention* information are computed.

Fig. 4 visualizes the results. It reveals that the objects which are described second and third are on average larger than those which are described later on. The objects which are more attention-worthy are on average closer to the beginning of the sentence for both human-written and machine-generated descriptions. Overall, there exists a similar trend between the captioning methods and humans, in which the attention decreases as getting away from the beginning. Looking into individual objects, some class categories may not follow this trend, e.g., 'person'. To conclude, despite small differences, both human and machine try to address the attention worthy objects as close to the beginning of a sentence as possible.

**Attention on described objects.** We signify the role of attention in descriptions by measuring the attention on described objects. We follow the steps of [64] and extend to machine-generated descriptions to compare descriptions by human and machine. We compute the probability of an object being fixated, $f$, given it is described, $d$, and visually exists, $e$, denoted as $p(f|d,e)$, and the probability of an object being described given it is fixated and exists, $p(d|f,e)$. We also computed the probability of an object being described given it visually exists, $p(d|e)$. Another interesting statistics is the probability of referrals to visually absent objects, i.e., $p(d|\neg e)$. The results are summarized in Table 1.

We learn that the human and the "Microsoft" model perform above the chance level, showing a correlation between attention and description in terms of $p(f|d,e)$. The low $p(d|f,e)$ is in agreement with the expectation of people looking around in order to describe something rather than describing something and looking around. We also learned that human describes existing objects more often compared to a machine and makes less referrals to non-existing objects. Intuitively, the small value of $p(d|\neg e)$ for human can be due to the use of nouns referring to concepts, scene schemes or an implicit piece of information.

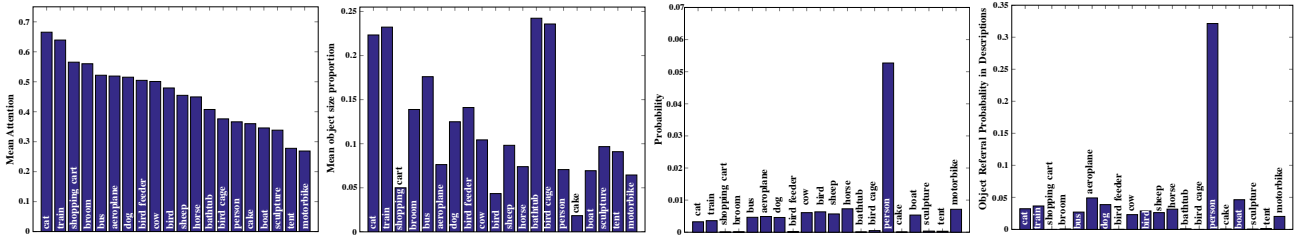

Figure 3. Analyzing human sentences and object class categories, from left to right, the top 20 attended objects in images, their normalized size, the probability of occurrence of the top 20 attended objects in an image, and their occurrence probability in descriptions.

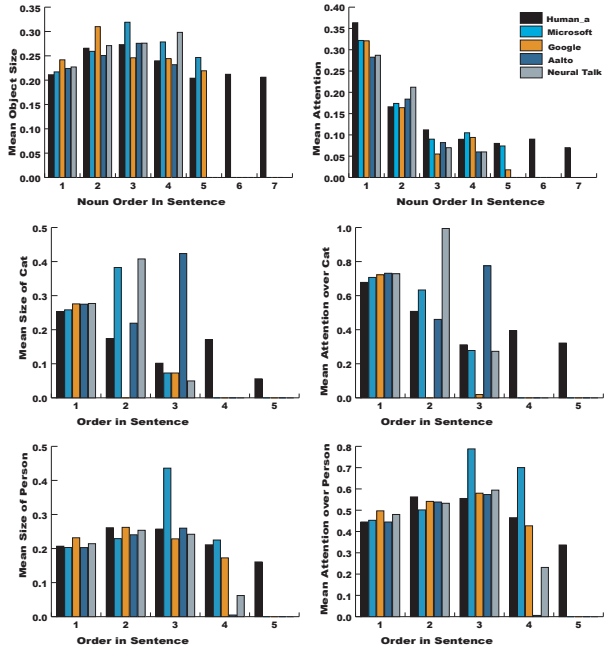

Figure 4. Object importance and referral order, from top to bottom: average over all object class categories, specific class categories: 'cat' and 'person'.

| Model | $p(f\|d,e)$ | $p(d\|f,e)$ | $p(d\|e)$ | $p(d\|\neg e)$ |
|---|---|---|---|---|
| Human | 0.6022 | 0.3023 | 0.2115 | 0.0939 |
| Microsoft | 0.6022 | 0.3028 | 0.2064 | 0.1017 |
| Google | 0.5273 | 0.2920 | 0.1945 | 0.1181 |
| Aalto | 0.5359 | 0.2949 | 0.2041 | 0.1052 |
| Neural Talk | 0.5273 | 0.2925 | 0.1872 | 0.1326 |

Table 1. Statistics of attention and described objects in descriptions by human and machine.

For the sake of reproducibility, we here elaborate the small details of this computation. There are cases that a description refers to a visually existing object multiple times, e.g., "a man and a boy ..." because an image may contain multiple objects of the same class category (in this example 'person'). In such cases, we account the referral only once and consider it fixated if any of the corresponding hand-labelled masks of that class category is fixated. The grounding between nouns and masks is validated and established by VOS.

To explain lower $p(f|d,e) = 0.60$ compared to the reported $p(f|d,e) = 0.87$ in [64], it is worth noting that, in this study, the object categories are obtained from contextual annotation, consisting of 222 classes compared to 20 in [64], and do not discriminate the background from the foreground objects. Also, a substantially higher number of descriptions are used to compute the human performance.

**The attention agreement between human and machine.** We quantified attention agreement by generating saliency maps from descriptions and checking their consistency with human attention on images. Given a description, we fetched all the referred objects and assigned them an attention value depending on their referral order. The attention value is obtained empirically from the average of human attention on object's referral order as in Fig 4. We put more weight to the centers of objects, as the object center is shown to allocate more fixations [7], and slightly smooth the maps. Some generated saliency map examples are provided in Fig. 5 for the captions given in Fig. 1.

Having a saliency map and fixation information, we employ the trivial fixation prediction evaluation criteria [8] for assessing a sentence in terms of attention. The average score over sentences of a model indicates the model's mean similarity with human in terms of attention. We utilized *area under the curve* (AUC) [30], *correlation coefficient* (CC), and *normalized scanpath saliency* (NSS) [47].

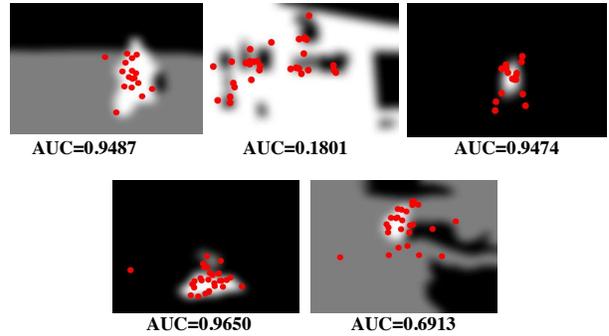

Figure 5. Example saliency maps generated from the sentences given in Figure 1 with the fixations overlaid.

|              | **Performance metrics** |        |        |
|--------------|:----:|:------:|:------:|
| **Model**    | AUC  | CC     | NSS    |
| Human        | 0.6811 | 0.1358 | 0.8008 |
| Microsoft    | 0.6707 | 0.1277 | 0.7561 |
| Google       | 0.6644 | 0.1198 | 0.7078 |
| Aalto        | 0.6701 | 0.1233 | 0.7235 |
| Neural Talk  | 0.6663 | 0.1229 | 0.7196 |

Table 2. Model performance in terms of attention: the evaluation of saliency maps generated from descriptions using fixations.

|              | **Performance metrics** | | | | |
|--------------|:-----:|:-----:|:------:|:------:|:-----:|
| **Model**    | AUC   | CIDEr | METEOR | ROUGEL | BLEU4 |
| **Microsoft**| **0.671** | **0.664** | **0.316** | **0.650** | **0.430** |
| **Aalto**    | 0.670 | 0.616 | 0.298  | 0.634  | **0.430** |
| **Google**   | 0.664 | 0.524 | 0.273  | 0.602  | 0.330 |
| **Neural Talk** | 0.666 | 0.503 | 0.273 | 0.593 | 0.323 |

Table 3. Performance of captioning models on augmented PASCAL50S database, traditional metrics and AUC are reported. The models are sorted based on AUC.

For all these measures larger values indicate better performance. To obtain an upper bound, we generated saliency maps for human-written descriptions and assessed their agreement with attention.

The results are summarized in Table 2. It is not surprising that the human-written sentences have the highest agreement with attention. However, there are cases in which the attention does not agree well for human-written captions. For example, consider the second image in Fig. 1 and its corresponding saliency in Fig. 5. As can be observed the human-description do not contain an explicit reference to the attended objects, but refer to a general concept.

**Comparing attention score with description scores.** From Table 2, we learn that the captioning methods differ with each other in terms of attention agreement with human. This motivates to gain further insight about the overall goodness of a model and its attention agreement with human. We thus evaluate the generated descriptions using Microsoft COCO caption evaluation code [9] and compared it with the AUC score of models. The results are summarized in Table 3. The results indicate that (1) the average ranking of methods by AUC agrees with the traditional metrics, and (2) the better a model is, the better attention agreement it has with human.

## 6. Saliency-Boosted Captioning Model

To this point, we confirmed that there exists a degree of agreement between descriptions by human and machine in terms of attention. We learned that better captioning models have a higher attention agreement with human. For this purpose, we relied on fixations gathered from a free-viewing task. Thus, we build a captioning model with visual features boosted by a saliency model in order to investigate potential improvements using a bottom-up saliency model. In other words, we focus on answering: *Can saliency benefit image captioning by machine*?

For this purpose, we employ a standard captioning model, based on an LSTM network of three layers with residual connections [24] between the layers. We use the open implementation of [53], where we set both feature input channels of the LSTM model to visual features and avoid any contextual features for simplicity. Fig. 6 depicts a high-level illustration of the proposed captioning model. In the following paragraphs, we explain the feature extraction, saliency computation and feature boosting and linearization. We refer the readers to [53] for the details of the language model.

**Image features.** We extract the image features using CNN features of the VGG network [54]. We follow the filter bank approach of [10] and compute the responses over the input image. In other words, the output of the last convolutional layer (*pool5*) is used. This results in a feature tensor of $7 \times 7 \times 512$, i.e., a $7 \times 7$ map of 512-dimensional feature vectors. These features are later boosted and linearized in order to be fed to the LSTM module.

**Saliency.** We compute the saliency using the image features of VGG network in order to be consistent with the image feature pipeline. Afterwards, we learn a regression to approximate the human fixations using extreme learning machines [27], following the saliency model of [59]. That is, an ensemble of saliency predictors are learnt to perform a regression from image features to the saliency space. The final saliency is the mean of the predicted saliencies from the members of the ensemble. The model with the saliency from VGG features will be called "**Proposed (VGG)**".

The VGG features are not fine-tuned for the specific task of saliency prediction, and are treated as generic descriptors [3], preventing the explicit learning of top-down factors that contribute to saliency prediction task, i.e., the regions that attract gaze such as faces. The saliency computation is hence bottom-up and in the category of learning-based saliency models [66].

Despite the proposed model of saliency computation is bottom-up, to prevent arguments on the role of implicitly learned top-down factors, we also used the saliency maps from a traditional pure bottom-up model, that is the maps from Graph-Based Visual Saliency (GBVS) [23] are used in the pipeline. We will refer to the model with GBVS saliency maps as "**Proposed (GBVS)**".

**Boosting image features and linearization.** We boost the CNN features before feeding them to the language model with the saliency map of the image. The procedure is depicted in Fig 7.

The CNN feature maps and saliency maps are of size $7 \times 7$. The saliency maps are normalized so that each pixel has a value between 0 and 1. We apply a $3 \times 3$ moving window with stride 1 on the $7 \times 7$ *pool5* feature map and concatenate the features under the $3 \times 3$ window to a feature vector of 4608 dimensions, denoted as $F_l$. A mean pooling is also performed on the $7 \times 7$ saliency map with a $3 \times 3$ moving window, denoted as $Sal$. Thus, a $5 \times 5$ feature map of 4608-dimensional feature vectors and a $5 \times 5$ aggregated saliency map are combined to produce a feature vector input to the language model.

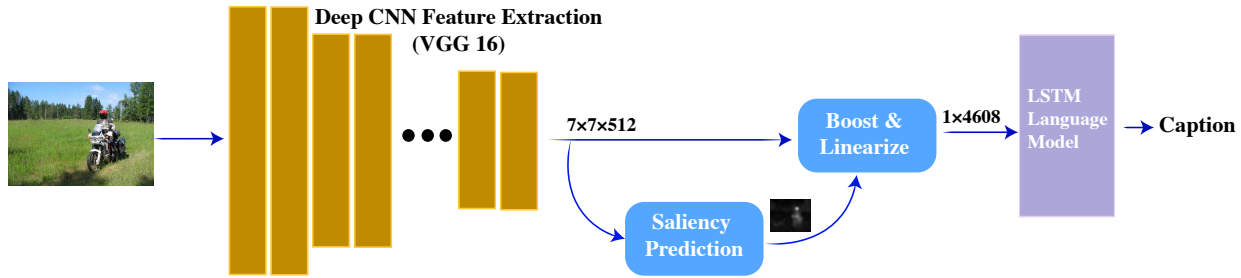

Figure 6. A high level outline of the proposed saliency-boosted captioning model.

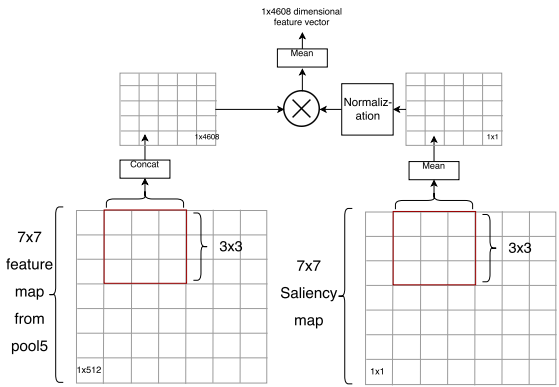

Figure 7. Block diagram showing the saliency boosting and feature linearization before feeding the feature to the language model.

To combine this saliency data with the image features, the local features corresponding to the feature maps are weighted by their corresponding saliency value. Then, the weighted feature vectors are averaged to produce a single feature vector, $F_{sal}$, which is input to the language model:

$$F_{sal} = \sum_{i=1}^{7} \sum_{j=1}^{7} Sal_\alpha(i,j) F_l(i,j), \quad (2)$$

$$Sal_\alpha(i,j) = 1 + Sal_{L_1}(i,j)^\alpha, \quad (3)$$

where $F_l$ is the image feature map, $Sal_{L_1}$ is the $L_1$ normalized saliency map, and $\alpha$ is an attenuation factor to control compactness of the aggregated saliency. The value of $\alpha = 2$ was determined via cross validation during training on MS COCO.

**Evaluating saliency contribution.** We evaluate the saliency boosted model on MS COCO [36] and augmented PASCAL50S, where the model is trained on MS COCO. To understand the contribution of saliency, we define a baseline using a uniform saliency where the saliency map is all ones. Then, we compare the proposed model and baseline. For the sake of completeness, we also include the performance of "Neural Talk" [32], "Google" [61], "Microsoft" [18], and "Aalto" [52].

|  | Performance metrics | | | |
|---|---|---|---|---|
| **Model** | CIDEr | METEOR | ROUGEL | BLEU4 |
| **Proposed (GBVS)** | 0.841 | 0.235 | <u>0.512</u> | <u>0.287</u> |
| **Proposed (VGG)** | 0.836 | 0.235 | 0.508 | 0.283 |
| **Prop. baseline** | 0.837 | 0.234 | 0.508 | 0.283 |
| **Microsoft** [18] | – | 0.236 | – | 0.257 |
| **Aalto** [52] | **0.899** | **0.243** | **0.520** | **0.299** |
| **Google** [61] | <u>0.855</u> | 0.237 | – | 0.277 |
| **Neural Talk** [32] | 0.660 | 0.195 | – | 0.230 |

Table 4. Performance on the COCO evaluation set, according to reported results.

The results on the MS COCO evaluation set are reported in Table 4. Comparing the proposed baseline and the saliency-boosted model, there is no significant improvement by boosting the model using a saliency in this dataset.

The results on the augmented PASCAL50S are reported in Table 5. As depicted, contrary to the MS COCO results, the proposed model outperforms the baseline, indicating that a better generalization is achieved by using saliency boosting considering both saliency models. We should also note here that the automatic evaluation is much more reliable on the PASCAL50S dataset owing to the availability of 50 references per image.

Overall, comparing between the results obtained on the MS COCO evaluation and augmented PASCAL50S datasets, the saliency boosting seems not contributing when we are training and testing a model on one database. The reason most likely lies on the fact that the model learns the underlying data very well and there is no need for further boosting. The performance results, however, shows that for across database and an unseen data with different visual characteristics, the saliency boosting contributes and improves the performance of the captioning model.

|  | **Performance metrics** | | | |
|---|---|---|---|---|
| **Model** | CIDEr | METEOR | ROUGEL | BLEU4 |
| **Proposed (GBVS)** | 0.615 | 0.302 | 0.635 | 0.400 |
| **Proposed (VGG)** | 0.614 | 0.302 | 0.632 | 0.396 |
| **Prop. baseline** | 0.585 | 0.293 | 0.624 | 0.375 |
| **Microsoft** | 0.664 | 0.316 | 0.650 | 0.430 |
| **Aalto** | 0.616 | 0.298 | 0.634 | 0.430 |
| **Google** | 0.524 | 0.273 | 0.602 | 0.330 |
| **Neural Talk** | 0.503 | 0.273 | 0.593 | 0.323 |

Table 5. Performance of captioning models on augmented Pascal50S, sorted by CIDEr.

## 7. Discussion and Conclusion

(1) *How well do image descriptions, by humans or models, on a scene agree with saliency*? In summary, all the captioning models show a clear degree of agreement with human in capturing saliency.

We testified that humans describe fixated items rather than looking aimlessly, consistent with [64]. Then, we extended the analysis to include descriptions by machine. The "Microsoft" model (the best among all models in this study) has the highest degree of agreement with human in terms of captioning metrics and attention. This indicates that the captioning models are becoming powerful enough to describe the visually existing elements similar to humans.

Coinciding with the supporting evidence about the role of saliency, e.g. [64, 28], we confirmed that more salient objects appear on average closer to the beginning of the descriptions by human. Extending the analysis to automatic captioning models, we observed a similar phenomenon. A model, superior to its counterparts, agrees with humans better in terms of attention on the order of nouns.

We quantified the attention agreement between human and machine by generating saliency maps from descriptions. The analysis shows that all the captioning models have a degree of agreement in terms of attention. The descriptions by "Microsoft" and "Aalto" models, which employ some level of object detection, are in the highest degree of agreement with human in terms of attention. This promotes possible contributions from top-down attention and contextual factors.

(2) *Can saliency benefit image captioning by machine*? The answer is not straightforward. We investigated the role of bottom-up saliency by proposing a simple saliency-boosted captioning model. The use of bottom-up attention coincides with the role of saliency in language formation, which followers of Wundt's theory mostly back. We learned that the saliency-boosted model is not significantly different from the baseline on the validation set of the MS COCO database, which the model was trained on. But, more importantly, the saliency-boosted model has a better generalization performance on augmented PASCAL50S dataset, which is totally unseen during the training, compared to the baseline.

(3) *What can we learn about top-down attention*? To infer any conclusion about the top-down mechanism of attention in sentence constructs, we require a data with simultaneous recording of eye movements and scene descriptions. In other words, given the limitation of the current data, which is based on free-viewing gaze, any concrete conclusion about top-down attention and models which are based on such an attention-mechanism, e.g. soft-attention [62], is impossible. We thus left out the top-down attention and soft-attention variants for future studies when simultaneous recording of eye movements and descriptions is available.

(4) *Can this study benefit from large scale click-based data*? Recent advancements based on deep-learning and the necessity of large databases for such approaches have enforced the community for seeking solutions like crowd-sourced databases using alternative mediums such as mouse clicks and mouse movements. In saliency research, SALICON [29] is the largest crowd-source based database using mouse movements. It includes a subset of MS COCO [36] images and mouse movements from participants in a free-viewing mouse movement paradigm. There is, however, existing research showing that while mouse movements may provide a first order approximation to eye tracking, it falls short on grasping the contextual information on a fine-grained manner and is recommended to be avoided for interpreting saliency results [58]. Following evaluation purposes for understanding saliency contribution, we, thus, avoid using data that relies on mouse movements.

(5) *Comments on emerging similar studies.* At the time of this writing, there exists several similar studies based on attention contribution in language modeling. Ramanishka et al. [49] create a model for inferring top-down attention from captions. Sugano and Bulling [57] explore the possibility of building a gaze-assisted captioning system using SALICON data. Pedersoli et al. [46] extend the soft-attention mechanism by regressing the location of objects for feature selection. In contrast, apart from technical differences, that is (1) being bottom-up, (2) employing human gaze, we have a philosophical difference in motive. While most of such works are following the research track of improving a captioning model or developing a system for assistive purposes, we look into the fundamental questions of the relation between saliency and sentence constructs and how saliency may benefit captioning. Our results, however, benefit the design of captioning models like [53] which use saliency.

**Acknowledgements.** The support of the Finnish Center of Excellence in Computational Inference Research and generous gifts from NVIDIA are gratefully acknowledged.

## References

[1] Microsoft COCO image captioning challenge. http://competitions.codalab.org/competitions/3221. Accessed: 2016-03-01.

[2] S. Antol, A. Agrawal, J. Lu, M. Mitchell, D. Batra, C. L. Zitnick, and D. Parikh. VQA: Visual question answering. In *ICCV*, 2015.

[3] H. Azizpour, A. S. Razavian, J. Sullivan, A. Maki, and S. Carlsson. From generic to specific deep representation for visual recognition. In *CVPR Workshops*, 2015.

[4] A. C. Berg, T. L. Berg, H. Daum, J. Dodge, A. Goyal, X. Han, A. Mensch, M. Mitchell, A. Sood, K. Stratos, and K. Yamaguchi. Understanding and predicting importance in images. In *CVPR*, 2012.

[5] R. Bernardi, R. Cakici, D. Elliott, A. Erdem, E. Erdem, N. Ikizler-Cinbis, F. Keller, A. Muscat, and B. Plank. Automatic description generation from images: A survey. *J. Artif. Intell. Res.*, 55(1), 2016.

[6] K. Bock, D. Irwin, D. Davidson, and W. Levelt. Minding the clock. *J. Mem. Lang.*, 48, 2003.

[7] A. Borji and J. Tanner. Reconciling saliency and object center-bias hypotheses in explaining free-viewing fixations. *IEEE Trans Neural Netw Learn Syst.*, 27(6), 2016.

[8] A. Borji, H. R. Tavakoli, D. N. Sihite, and L. Itti. Analysis of scores, datasets, and models in visual saliency prediction. In *ICCV*, 2013.

[9] X. Chen, T.-Y. L. Hao Fang, R. Vedantam, S. Gupta, P. Dollr, and C. L. Zitnick. Microsoft COCO captions: Data collection and evaluation server. *CoRR*, abs/1504.00325, 2015.

[10] M. Cimpoi, S. Maji, and A. Vedaldi. Deep filter banks for texture recognition and segmentation. In *CVPR*, 2015.

[11] A. D. F. Clarke, M. Elsner, and H. Rohde. Giving good directions: order of mention reflects visual salience. *Front. Psychol.*, 6, 2015.

[12] M. Denkowski and A. Lavie. Meteor universal: Language specific translation evaluation for any target language. In *EACL*, 2014.

[13] J. Devlin, H. Cheng, H. Fang, S. Gupta, L. Deng, X. He, G. Zweig, and M. Mitchell. Language models for image captioning: The quirks and what work. In *ACL*, 2015.

[14] J. Donahue, L. Hendricks, S. Guadarrama, M. Rohrbach, S. Venugopalan, K. Saenko, and T. Darrell. Long-term recurrent convolutional networks for visual recognition and description. In *CVPR*, 2015.

[15] D. Elliott and A. P. de Vries. Describing images using inferred visual dependency representations. In *ACL*, 2015.

[16] D. Elliott and F. Keller. Image description using visual dependency representations. In *EMNLP*, 2013.

[17] M. Everingham, S. M. A. Eslami, L. Van Gool, C. K. I. Williams, J. Winn, and A. Zisserman. The pascal visual object classes challenge: A retrospective. *IJCV*, 111(1), 2015.

[18] H. Fang, S. Gupta, F. Iandola, R. Srivastava, L. Deng, P. Dollar, J. Gao, X. He, M. Mitchell, J. Platt, L. Zitnick, and G. Zweig. From captions to visual concepts and back. In *CVPR*, 2015.

[19] A. Farhadi, M. Hejrati, M. A. Sadeghi, P. Young, C. Rashtchian, J. Hockenmaier, and D. Forsyth. Every picture tells a story: Generating sentences from images. In *ECCV*, 2010.

[20] M. R. Greene. Statistics of high-level scene context. *Front. Psychol.*, 4, 2013.

[21] Z. Griffin and K. Bock. What the eyes say about speaking. *Psychol Sci.*, 11(4), 2000.

[22] Z. M. Griffin and D. H. Spieler. Observing the what and when of language production for different age groups by monitoring speakers eye movements. *Brain and Language*, 99(3):272 – 288, 2006. Language Comprehension across the Life Span.

[23] J. Harel, C. Koch, and P. Perona. Graph-based visual saliency. In *NIPS*, 2006.

[24] K. He, X. Zhang, S. Ren, and J. Sun. Deep residual learning for image recognition. In *CVPR*, 2016.

[25] M. Hodosh, P. Young, and J. Hockenmaier. Framing image description as a ranking task: Data, models and evaluation metrics. *J. Artif. Intell. Res.*, 47, 2013.

[26] J. Holsanova. *How we focus attention in picture viewing, picture description, and during mental imagery*, pages 291–313. Bilder - sehen - denken : zum Verhltnis von begrifflich-philosophischen und empirisch-psychologischen Anstzen in der bildwissenschaftlichen Forschung. von Halem, 2011.

[27] G.-B. Huang, Q.-Y. Zhu, and C.-K. Siew. Extreme learning machine: Theory and applicatons. *Neurocomput.*, 70, 2006.

[28] L. Itti and M. A. Arbib. *Action to Language via the Mirror Neuron System*, chapter Attention and the Minimal Subscene. Combridge Press, 2006.

[29] M. Jiang, S. Huang, J. Duan, and Q. Zhao. SALICON: Saliency in context. In *CVPR*, 2015.

[30] T. Judd, K. Ehinger, F. Durand, and A. Torralba. Learning to predict where humans look. In *ICCV*, 2009.

[31] M. k. Tanenhaus, C. Chambers, and J. E. Hanna. Referential domains in spoken language comprehension: Using eye movements to bridge the product and action traditions. In *The interface of language, vision, and action: Eye movements and visual world*. Psychology Press, 2004.

[32] A. Karpathy and L. Fei-Fei. Deep visual-semantic alignments for generating image descriptions. In *CVPR*, 2015.

[33] A. Karpathy, A. Joulin, and L. Fei-Fei. Deep fragment embeddings for bidirectional image sentence mapping. In *NIPS*, 2014.

[34] S. Li, G. Kulkarni, T. L. Berg, A. C. Berg, and Y. Choi. Composing simple image descriptions using web-scale n-grams. In *CoNLL*, 2011.

[35] C.-Y. Lin. Rouge: A package for automatic evaluation of summaries. In *Text Summarization Branches Out: Proceedings of the ACL Workshop*, 2004.

[36] T.-Y. Lin, M. Maire, S. Belongie, J. Hays, P. Perona, D. Ramanan, P. Dollar, and C. L. Zitnick. Microsoft COCO: Common objects in context. In *ECCV*, 2014.

[37] M. Malinowski, M. Rohrbach, and M. Fritz. Ask your neurons: A neural-based approach to answering questions about images. In *ICCV*, 2015.

[38] C. D. Manning, M. Surdeanu, J. Bauer, J. Finkel, S. J. Bethard, and D. McClosky. The Stanford CoreNLP natural language processing toolkit. In *ACL*, 2014.

[39] S. Mathe and C. Sminchisescu. Actions in the eye: Dynamic gaze datasets and learnt saliency models for visual recognition. *IEEE Trans. Pattern Anal. Mach. Intell.*, 2015.

[40] A. S. Meyer. The use of eye tracking in studies of sentence generation. In *The interface of language, vision, and action: Eye movements and the visual world*. Psychology Press, 2004.

[41] A. S. Meyer, A. M. Sleiderink, and W. J. Levelt. Viewing and naming objects: eye movements during noun phrase production. *Cognition*, 66(2), 1998.

[42] T. Mikolov, K. Chen, G. Corrado, and J. Dean. Efficient estimation of word representations in vector space. *ICLR*, 2013.

[43] R. Mottaghi, X. Chen, X. Liu, N. G. Cho, S. W. Lee, S. Fidler, R. Urtasun, and A. Yuille. The role of context for object detection and semantic segmentation in the wild. In *CVPR*, 2014.

[44] V. Ordonez, G. Kulkarni, and T. L. Berg. Im2text: Describing images using 1 million captioned photographs. In *NIPS*, 2011.

[45] K. Papineni, S. Roukos, T. Ward, and W. jing Zhu. Bleu: a method for automatic evaluation of machine translation. In *ACL*, 2002.

[46] M. Pedersoli, T. Lucas, C. Schmid, and J. Verbeek. Areas of attention for image captioning. In *CVPR*, 2017.

[47] R. J. Peters, A. Iyer, L. Itti, and C. Koch. Components of bottom-up gaze allocation in natural images. *Vision Research*, 45, 2005.

[48] F. Pulvermüller, M. Hrle, and F. Hummel. Walking or talking?: Behavioral and neurophysiological correlates of action verb processing. *Brain and Language*, 78(2), 2001.

[49] V. Ramanishka, A. Das, J. Zhang, and K. Saenko. Top-down visual saliency guided by captions. In *CVPR*, 2017.

[50] C. Rashtchian, P. Young, M. Hodosh, and J. Hockenmaier. Collecting image annotations using amazon's mechanical turk. In *NAACL HLT*, 2010.

[51] A. Rohrbach, M. Rohrbach, and B. Schiele. The Long-Short Story of Movie Description. In *GCPR*, 2015.

[52] R. Shetty and J. Laaksonen. Video captioning with recurrent networks based on frame- and video-level features and visual content classification. In *CVPR Workshops*, 2015.

[53] R. Shetty, H. R-Tavakoli, and J. Laaksonen. Exploiting scene context for image captioning. In *ACMMM Vision and Language Integration Meets Multimedia Fusion Workshop*, 2016.

[54] K. Simonyan and A. Zisserman. Very deep convolutional networks for large-scale image recognition. *CoRR*, abs/1409.1556, 2014.

[55] D. D. Sleator and D. Temperley. Parsing english with a link grammar. In *Third International Workshop on Parsing Technologies*, 1991.

[56] S. N. Sridhar. *Cognition and Sentence Production: A Cross-Linguistic Study*, chapter Models of Sentence Production, pages 7–19. Springer New York, 1988.

[57] Y. Sugano and A. Bulling. Seeing with humans: Gaze-assisted neural image captioning. *CoRR*, abs/1608.05203, 2016.

[58] H. R. Tavakoli, F. Ahmad, A. Borji, and J. Laaksonen. Saliency revisited: Analysis of mouse movements versus fixations. In *CVPR*, 2017.

[59] H. R. Tavakoli, A. Borji, J. Laaksonen, and E. Rahtu. Exploiting inter-image similarity and ensemble of extreme learners for fixation prediction using deep features. *Neurocomput.*, 244, 2017.

[60] R. Vedantam, C. L. Zitnick, and D. Parikh. CIDEr: Consensus-based image description evaluation. In *CVPR*, 2015.

[61] O. Vinyals, A. Toshev, S. Bengio, and D. Erhan. Show and tell: A neural image caption generator. In *CVPR*, 2015.

[62] K. Xu, J. Ba, R. Kiros, K. Cho, A. Courville, R. Salakhudinov, R. Zemel, and Y. Bengio. Show, attend and tell: Neural image caption generation with visual attention. In F. Bach and D. Blei, editors, *Proceedings of the 32nd International Conference on Machine Learning*, volume 37 of *Proceedings of Machine Learning Research*, pages 2048–2057, Lille, France, 07–09 Jul 2015. PMLR.

[63] P. Young, A. Lai, M. Hodosh, and J. Hockenmaier. From image descriptions to visual denotations: New similarity metrics for semantic inference over event descriptions. *TACL*, 2, 2014.

[64] K. Yun, Y. Peng, D. Samaras, G. Zelinsky, and T. Berg. Exploring the role of gaze behavior and object detection in scene understanding. *Front. Psychol.*, 4, 2013.

[65] K. Yun, Y. Peng, D. Samaras, G. J. Zelinsky, and T. L. Berg. Studying relationships between human gaze, description, and computer vision. In *CVPR*, 2013.

[66] Q. Zhao and C. Koch. Learning saliency-based visual attention: A review. *Signal Processing*, 93, 2013.

[67] C. L. Zitnick, R. Vedantam, and D. Parikh. Adopting abstract images for semantic scene understanding. *IEEE Trans. Pattern Anal. Mach. Intell.*, 38(4), 2016.

[68] B. M. t Hart, H. C. E. F. Schmidt, C. Roth, and W. Einhauser. Fixations on objects in natural scenes: dissociating importance from salience. *Front. Psychol.*, 4, 2013.